\pdfoutput=1

\documentclass{article}

\PassOptionsToPackage{numbers, compress}{natbib}

\usepackage[preprint]{neurips_2025}

\usepackage[utf8]{inputenc} 
\usepackage[T1]{fontenc}    
\usepackage{url}            
\usepackage{booktabs}       
\usepackage{amsmath,amsfonts,amssymb}       
\usepackage{nicefrac}       
\usepackage{microtype}      
\usepackage{xcolor}         
\usepackage{float}
\usepackage{graphicx}

\usepackage{booktabs} 
\usepackage{array}    

\usepackage{algorithm}
\usepackage{algpseudocode}
\usepackage{wrapfig}
\usepackage{float} 
\usepackage{setspace}       

\usepackage{microtype}
\usepackage{graphicx}
\usepackage{subcaption}
\usepackage{adjustbox}
\usepackage{booktabs} 
\usepackage{chngcntr}
\usepackage{changepage}

\usepackage[colorlinks=true, allcolors=blue]{hyperref}
\usepackage[capitalize, noabbrev, nameinlink]{cleveref}

\usepackage{enumitem}

\usepackage{amsmath}
\usepackage{amssymb}
\usepackage{mathtools}
\usepackage{amsthm}

\usepackage{math_commands}
\usepackage{newtxmath}

\makeatletter

\renewcommand{\appendixautorefname}{\S\@gobble}
\renewcommand{\sectionautorefname}{\S\@gobble}
\renewcommand{\subsectionautorefname}{\S\@gobble}
\renewcommand{\subsubsectionautorefname}{\S\@gobble}

\makeatother

\makeatletter

\providecommand{\section}{}
\renewcommand{\section}{%
  \@startsection{section}{1}{\z@}%
                {-1.0ex \@plus -0.2ex \@minus -0.2ex}%
                { 1.0ex \@plus  0.2ex \@minus  0.2ex}%
                {\large\bf\raggedright}%
}
\providecommand{\subsection}{}
\renewcommand{\subsection}{%
  \@startsection{subsection}{2}{\z@}%
                {-1.0ex \@plus -0.2ex \@minus -0.2ex}%
                { 0.3ex \@plus  0.2ex}%
                {\normalsize\bf\raggedright}%
}
\providecommand{\subsubsection}{}
\renewcommand{\subsubsection}{%
  \@startsection{subsubsection}{3}{\z@}%
                {-1.0ex \@plus -0.2ex \@minus -0.2ex}%
                { 0.3ex \@plus  0.2ex}%
                {\normalsize\bf\raggedright}%
}
\providecommand{\paragraph}{}
\renewcommand{\paragraph}[1]{\textbf{#1}}
\providecommand{\subparagraph}{}
\renewcommand{\subparagraph}{%
  \@startsection{subparagraph}{5}{\z@}%
                {1.5ex \@plus 0.5ex \@minus 0.2ex}%
                {-1em}%
                {\normalsize\bf}%
}

\makeatother

\pdfoutput=1

\title{
LoRA Is Slower Than You Think}

\author{
Seokmin Ko$^{1}$\thanks{Correspondence to: \texttt{min11947@yonsei.ac.kr}}\\
\\
$^1$Yonsei University \\
}

\begin{document}

\maketitle
\begin{abstract}
\looseness=-1
Low-Rank Adaptation (LoRA) is one of the most widely used techniques for fine-tuning large language models (LLMs)\citep{Hu2022LoRA,dettmers2023qlora,li2021prefix}. By introducing a small number of trainable low-rank weight matrices, LoRA substantially reduces the number of parameters that need to be updated, offering significant advantages in memory consumption and computational efficiency compared to full fine-tuning. However, we observed that LoRA does not consistently provide speed improvements across all model architectures and training setups. Motivated by this inconsistency, we conduct a comprehensive analysis of LoRA's performance and investigate the underlying factors limiting its speedup. Based on our findings, we propose several methods for more efficient fine-tuning of LLMs. We empirically evaluate these methods and compare them to LoRA, demonstrating that our approach achieves comparable or superior performance while delivering more consistent training speed improvements. Our work offers valuable insights and practical guidelines for practitioners seeking to optimize LLM fine-tuning under resource constraints.

\end{abstract}

\section{Introduction} \label{sec:intro}
Over the past few years, large language models (LLMs) have shown remarkable advances, driven primarily by rapid increases in model size. For example, the GPT series has scaled from 1B parameters in GPT-2 to 175B in GPT-3 \citep{zhang2021commentary}, with performance improvements roughly proportional to model capacity. However, the sheer scale of these models poses a significant challenge for fine-tuning: updating billions of parameters demands enormous memory and computational resources.
To tackle this bottleneck, parameter-efficient fine-tuning (PEFT) approaches have emerged\citep{houlsby2019parameter,li2021prefix}, enabling researchers and practitioners to adapt LLMs without full weight updates. Among these methods, Low-Rank Adaptation (LoRA) is particularly popular due to its simplicity and effectiveness \citep{huan2025fine}. LoRA inserts low-rank weight matrices $A\in\mathbb{R}^{d\times r}$ and $B\in\mathbb{R}^{r\times d}$ into each layer, reducing the number of trainable parameters from $d^2$ to $2dr$ and thus significantly lowering memory usage and gradient computation cost\citep{zaken2021bitfit}. As we will show in Related Work, LoRA offers theoretical benefits in reducing both memory footprint and computational overhead \citep{cherniuk2023run}.

Despite these theoretical gains, our experiments reveal that LoRA can be slower than full fine-tuning on certain setups. Specifically, on an NVIDIA A100 GPU (\citep{choquette2021nvidia}) with batch size 4 and sequence lengths of 512 and 1024, LoRA training sometimes incurs longer forward and backward times compared to standard fine-tuning (see Table~\ref{tab:performance}).

\begin{table}[t]
  \caption{Comparison of training times (forward/backward) between LoRA and full fine‐tuning on various models using an A100 GPU (batch size 4, seq.\ lengths 512/1024).}
  \label{tab:performance}
  \centering
  \small
  \begin{tabular}{l|l|c|c}
    \hline
    Model                & Training Type & Forward Time (ms)     & Backward Time (ms)    \\
    \hline
    GPT2-xl (1.5B)       & LoRA          & 97.73 / 190.40        & 124.34 / 246.82       \\
                         & Full-FT       & 61.89 / 118.22        & 114.92 / 224.48       \\
    \hline
    GPT2-medium (345M)   & LoRA          & 48.68 / 62.17         & 44.66 / 77.11         \\
                         & Full-FT       & 27.50 / 38.08         & 36.12 / 69.32         \\
    \hline
    Tiny Llama (1.1B)    & LoRA          & 60.35 / 97.56         & 81.61 / 131.33        \\
                         & Full-FT       & 37.14 / 71.19         & 66.89 / 134.26        \\
    \hline
  \end{tabular}
\end{table}

Our findings motivate an investigation into the root causes of LoRA’s inconsistent speed gains and the development of more reliable adaptation techniques. The reason why such an unexpected phenomenon happens is because of the processing mechanism of GPU ~\citep{woo2025paca,wang2023non,dai2018accelerate,han2022microsecond}. Basically, GPUs are specialized in processing a kernel at once. By that property, the adaptive layers of LoRA are processed sequentially, which causes bottleneck. Even if LoRA has theoretical gains compared to Full Fine-Tuning in training phases, to benefit from LoRA in training speed requires huge amount of parameters to exceed the loss of LoRA by adpater's architecture.

To address the limitations identified in our study, we propose a \textit{selective non-adaptive fine-tuning} approach that omits low-rank adapter modules and selectively updates only the most task-critical parameters. We empirically demonstrate that this method achieves competitive or superior performance compared to LoRA, while substantially reducing training time.

Our contributions are as follows:
\begin{enumerate}
  \item We show that, contrary to conventional wisdom, LoRA does not always provide training speed advantages.
  \item We introduce a selective non-adaptive fine-tuning technique that delivers faster training speeds without sacrificing model performance.
\end{enumerate}

\section{Preliminaries}
\label{sec:prelim}

\subsection{Low-Rank Adaptation (LoRA)}
In their landmark paper, Hu et al. \citep{Hu2022LoRA} presented Low-Rank Adaptation (LoRA) as a parameter-efficient fine-tuning technique for large pretrained models. Rather than updating the entire weight matrix $W_0\in\mathbb{R}^{d_{out}\times d_{in}}$, LoRA represents the incremental update $\Delta W$ as the product of two low-rank matrices $A\in\mathbb{R}^{d_{out}\times r}$ and $B\in\mathbb{R}^{r\times d_{in}}$, where $r \ll \min(d_{in},d_{out})$:

\begin{equation}
W = W_0 + \Delta W, \quad \Delta W = BA.
\end{equation}

Only the matrices $A$ and $B$ are trained, reducing the number of tunable parameters from $d_{out}\times d_{in}$ to $r(d_{out}+d_{in})$, while keeping $W_0$ frozen. The authors demonstrated that LoRA with small ranks (e.g., $r=8$ or $16$) achieves performance on tasks like machine translation and question answering competitive with full fine-tuning, while significantly lowering memory usage and computational overhead.

We begin by formalizing the computational cost of full fine-tuning (Full-FT) and LoRA for a single layer with input dimension $d_{in}$, output dimension $d_{out}$, and batch size $N$. Let $W\in\mathbb{R}^{d_{out}\times d_{in}}$, $X_{in}\in\mathbb{R}^{d_{in}\times N}$, and $X_{out}\in\mathbb{R}^{d_{out}\times N}$ denote the weights of the layer, the input activations, and the output activations, respectively.

Full-FT performs the following operations:
\begin{align}
\text{Forward: } & X_{out} = W X_{in}, & O(N d_{out} d_{in}) \\
\text{Backward: } & \nabla X_{in} = W^T \nabla X_{out}, & O(N d_{out} d_{in}) \\
               & \nabla W = \nabla X_{out} X_{in}^T, & O(N d_{out} d_{in})
\end{align}
Full-FT thus incurs $O(3 N d_{out} d_{in})$ time complexity per layer, along with the corresponding memory overhead to store gradients and optimizer states.

LoRA updates only two low-rank matrices $A\in\mathbb{R}^{d_{out}\times r}$ and $B\in\mathbb{R}^{r\times d_{in}}$, with $r \ll \min(d_{in},d_{out})$:
\begin{align}
\text{Forward: } & X_{out} = W X_{in} + B\bigl(A X_{in}\bigr), & O(N d_{out} d_{in} + 2 N d r) \\
\text{Backward: } & \nabla X_{in} = W^T \nabla X_{out} + A^T \bigl(B^T \nabla X_{out}\bigr), & O(N d_{out} d_{in} + N d r) \\
               & \nabla B = \nabla X_{out} X_{mid}^T, ~~ \nabla A = \nabla X_{mid} X_{in}^T, & O(N d r)
\end{align}
LoRA reduces the complexity per layer to $O(2 N d_{out} d_{in} + 3 N d r)$, which is substantially lower when $r\ll d$.

\subsection{Partial Connection Adaptation (PaCA)}
\begin{figure}[ht]
  \centering
  \includegraphics[width=0.8\linewidth]{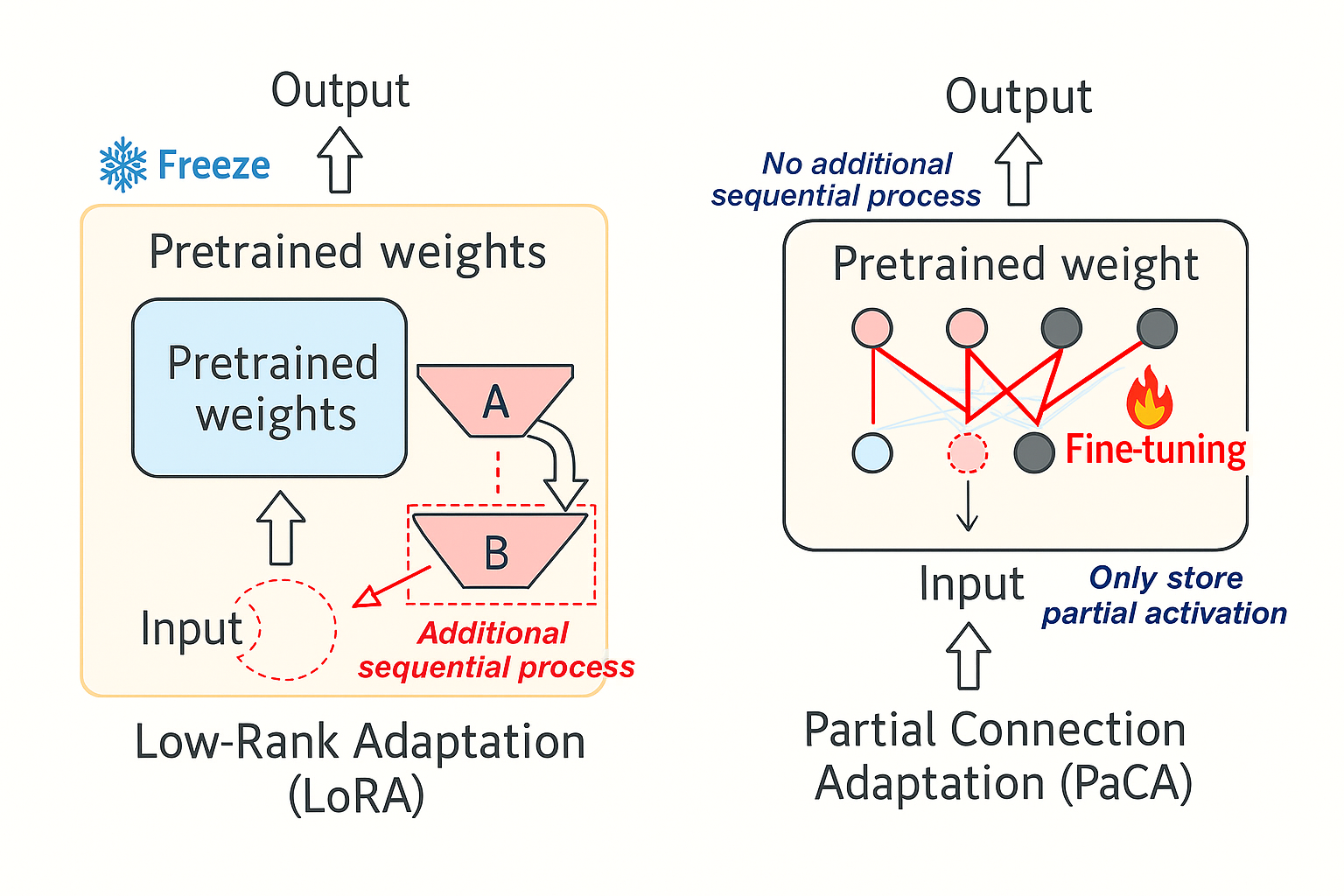}
  \caption{The Partial Connection Adaptation (PaCA) architecture}
  \label{fig:paca_arch}
\end{figure}

Woo et al. introduced Partial Connection Adaptation (PaCA), which further improves the efficiency of fine-tuning by selectively updating only a subset of weight connections. PaCA divides the weight matrix $W_0$ into blocks and employs a learnable binary mask $M\in\{0,1\}^{d_{out}\times d_{in}}$ to indicate which blocks should be updated. The update rule is defined as:

\begin{equation}
W = W_0 + M \odot \Delta W,
\end{equation}
\begin{align}
\text{Full-FT: } & W^{k+1} = W^{k} - \eta \nabla W^{k}, \\
\text{PaCA: } & W^{k+1} = W^{k} - \eta \bigl[M \odot \Delta W^{k}\bigr],
\end{align}

where $\odot$ denotes element-wise multiplication, and $\Delta W$ follows the same low-rank factorization as in LoRA. By learning the mask $M$, the model focuses parameter updates on the most critical connections. Figure~\ref{fig:paca_arch} illustrates the PaCA framework.

For a pretrained weight matrix of size 
4096
×
4096
, LoRA with rank 
=
8
 updates 
4096
×
16
 parameters. In PaCA, by selecting and updating only 16 columns (ranks) of the same matrix, we also update 
4096
×
16
 parameters—matching LoRA’s parameter count. The PaCA paper demonstrates that this selective column-wise fine-tuning achieves performance comparable to LoRA while delivering faster training speeds.

\section{Method}\label{sec:method}

In this work, we focus on Partial Connection Adaptation (PaCA) as a means to perform efficient fine‐tuning without introducing adapter modules. Rather than updating every layer of a Transformer, we propose a scheme \emph{selective non‐adaptive fine‐tuning} that only updates the upper layers of an LLM that are more task-relevant.

Transformer-based LLMs (e.g. LLaMA2 7B \citep{touvron2023llama}) consist of $L$ stacked layers (32 in the case of LLaMA2 7B).  In standard backpropagation (Hinton et al.~\citep{rumelhart1986learning} ), computing gradients for lower layers via the chain rule requires propagating errors sequentially through all layers, which incurs significant compute and memory overhead.  To break this bottleneck, we freeze the lower $L - K$ layers and apply PaCA only to the upper $K$ layers.  Concretely, if we choose $K = 16$ in a 32‐layer model, we selectively fine‐tune those 16 upper layers using PaCA.  To keep the total number of updated parameters identical to a full model LoRA of rank $r$, we double the PaCA mask density (that is, select columns $2r$ per layer rather than $r$).

\begin{figure}[ht]
  \centering
  \includegraphics[width=0.4\linewidth]{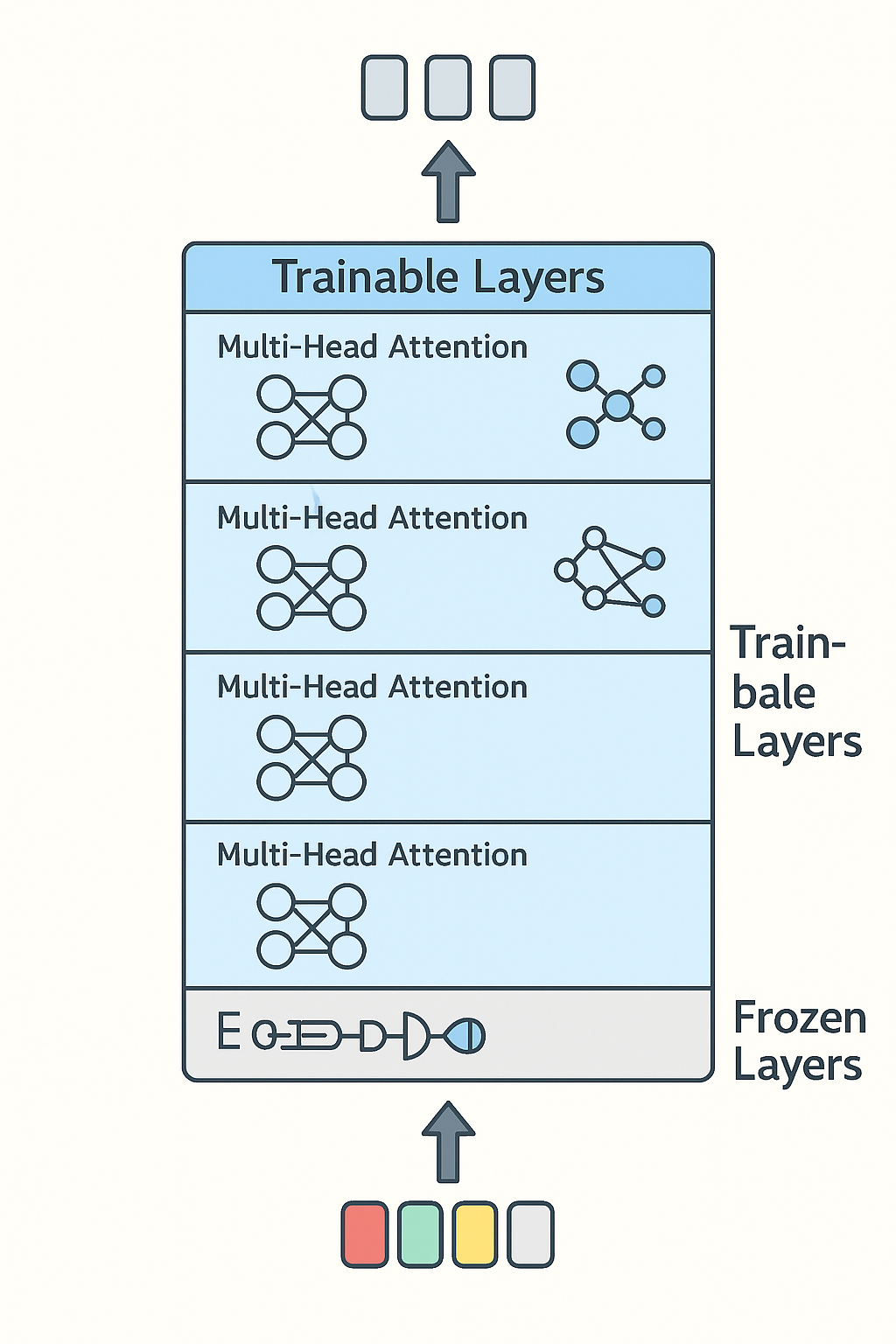}
  \caption{Selective non‐adaptive fine‐tuning: only the top $K$ layers of the pretrained model are updated via PaCA, while the remaining layers remain frozen.}
  \label{fig:selective_method}
\end{figure}

Figure~\ref{fig:selective_method} illustrates our selective non‐adaptive fine‐tuning pipeline.  By restricting gradient updates to a subset of layers and connections, we achieve faster wall clock training times while maintaining comparable performance.

\section{Experiments}\label{sec:experiments}

 We evaluate five tuning strategies—No tuning, LoRA, PaCA, Upper Half PaCA, and Three Quarters PaCA—on the MMLU benchmark using the LLaMA2-7B model\citep{hendrycks2020measuring}.  All experiments are run on a single NVIDIA A6000 GPU\citep{yang2024accurate} with batch size 8, gradient accumulation of 4, sequence length 512, and 1 epoch of training.  Table~\ref{tab:mmlu} reports the wall‐clock training times and category‐wise accuracies; Table~\ref{tab:hyperparams} lists the hyperparameter configurations for each method.

\begin{table}[t]
  \caption{MMLU Evaluation for LLaMA2-7B}
  \label{tab:mmlu}
  \centering
  \small
  \begin{tabular}{lcccccc}
    \toprule
    Method                & Training Time & Hums. & STEM  & Social. & Other & Avg.   \\
    \midrule
    No tuning             & –             & 44.0  & 37.0  & 51.5    & 53.1  & 45.9   \\
    LoRA                  & 7:17:01.79    & 50.0  & 42.2  & 59.7    & 58.0  & 52.15  \\
    PaCA                  & 6:04:29.65    & 49.5  & 41.8  & 60.4    & 58.1  & 52.04  \\
    Upper Half PaCA       & 4:19:17.76    & 44.8  & 38.4  & 52.7    & 53.9  & 47.10  \\
    Three Quarters PaCA   & 5:15:44.61    & 49.2  & 42.2  & 60.0    & 58.3  & 52.02  \\
    \bottomrule
  \end{tabular}
\end{table}

\begin{table}[t]
  \caption{Hyperparameters for LLaMA2-7B Methods}
  \label{tab:hyperparams}
  \centering
  \resizebox{0.9\linewidth}{!}{%
    \begin{tabular}{lcccc}
      \toprule
      Hyperparameter                  & LoRA     & Upper Half PaCA & Three Quarters PaCA & PaCA     \\
      \midrule
      Rank                            & 8        & 32              & 24                  & 16       \\
      $\alpha$                        & 32       & 64              & 64                  & 64       \\
      Dropout                         & 0.1      & –               & –                   & –        \\
      LR (LLaMA2-7B)                  & $3\times10^{-4}$ & $3\times10^{-4}/1\times10^{-4}$ & $3\times10^{-4}/1\times10^{-4}$ & $3\times10^{-4}/1\times10^{-4}$ \\
      Training precision              & 16-bit mixed & 16-bit mixed   & 16-bit mixed       & 16-bit mixed \\
      Optimizer                       & AdamW    & AdamW           & AdamW               & AdamW    \\
      LR scheduler                    & cosine   & cosine          & cosine              & cosine   \\
      Batch size                      & 8        & 8               & 8                   & 8        \\
      Gradient accumulation steps     & 4        & 4               & 4                   & 4        \\
      Sequence length                 & 512      & 512             & 512                 & 512      \\
      Warmup steps                    & 100      & 100             & 100                 & 100      \\
      Epochs                          & 1        & 1               & 1                   & 1        \\
      Target modules                  & Q, K, V, O, Up, Down, Gate & Q, K, V, O, Up, Down, Gate & Q, K, V, O, Up, Down, Gate & Q, K, V, O, Up, Down, Gate \\
      \bottomrule
    \end{tabular}%
  }
\end{table}

 As shown in Table \ref{tab:mmlu}, the Three Quarters PaCA model achieves performance comparable to both LoRA and full PaCA, while reducing training time by 2 hours and 2 minutes compared to LoRA and by 49 minutes compared to PaCA.
\section{Conclusion}

In this work, we showed that the widely used LoRA method, despite its theoretical advantages, does not always yield training‐time improvements on GPU hardware due to sequential adapter processing bottlenecks. To address this, we proposed a selective non-adaptive fine-tuning approach that forgoes adapter modules and updates only the most task-critical upper layers of the model. Our experiments on LLaMA2-7B with the MMLU benchmark demonstrate that this strategy preserves accuracy while reducing training time by up to 2 hours compared to LoRA and by 49 minutes compared to PaCA. Future work will explore automated layer-selection mechanisms and the extension of our method to other PEFT techniques.

\bibliography{reference}

\begin{thebibliography}{17}
\providecommand{\natexlab}[1]{#1}
\providecommand{\url}[1]{\texttt{#1}}
\expandafter\ifx\csname urlstyle\endcsname\relax
  \providecommand{\doi}[1]{doi: #1}\else
  \providecommand{\doi}{doi: \begingroup \urlstyle{rm}\Url}\fi

\bibitem[Cherniuk et~al.(2023)Cherniuk, Mikhalev, and Oseledets]{cherniuk2023run}
Daria Cherniuk, Aleksandr Mikhalev, and Ivan Oseledets.
\newblock Run lora run: Faster and lighter lora implementations.
\newblock \emph{arXiv preprint arXiv:2312.03415}, 2023.

\bibitem[Choquette et~al.(2021)Choquette, Gandhi, Giroux, Stam, and Krashinsky]{choquette2021nvidia}
Jack Choquette, Wishwesh Gandhi, Olivier Giroux, Nick Stam, and Ronny Krashinsky.
\newblock Nvidia a100 tensor core gpu: Performance and innovation.
\newblock \emph{IEEE Micro}, 41\penalty0 (2):\penalty0 29--35, 2021.

\bibitem[Dai et~al.(2018)Dai, Lin, Li, Zhao, Wang, Zheng, and Zhou]{dai2018accelerate}
Hongwen Dai, Zhen Lin, Chao Li, Chen Zhao, Fei Wang, Nanning Zheng, and Huiyang Zhou.
\newblock Accelerate gpu concurrent kernel execution by mitigating memory pipeline stalls.
\newblock In \emph{2018 IEEE international symposium on high performance computer architecture (HPCA)}, pages 208--220. IEEE, 2018.

\bibitem[Dettmers et~al.(2023)Dettmers, Pagnoni, Holtzman, and Zettlemoyer]{dettmers2023qlora}
Tim Dettmers, Artidoro Pagnoni, Ari Holtzman, and Luke Zettlemoyer.
\newblock Qlora: Efficient finetuning of quantized llms.
\newblock \emph{Advances in neural information processing systems}, 36:\penalty0 10088--10115, 2023.

\bibitem[Han et~al.(2022)Han, Zhang, Chen, and Chen]{han2022microsecond}
Mingcong Han, Hanze Zhang, Rong Chen, and Haibo Chen.
\newblock Microsecond-scale preemption for concurrent $\{$GPU-accelerated$\}$$\{$DNN$\}$ inferences.
\newblock In \emph{16th USENIX Symposium on Operating Systems Design and Implementation (OSDI 22)}, pages 539--558, 2022.

\bibitem[Hendrycks et~al.(2020)Hendrycks, Burns, Basart, Zou, Mazeika, Song, and Steinhardt]{hendrycks2020measuring}
Dan Hendrycks, Collin Burns, Steven Basart, Andy Zou, Mantas Mazeika, Dawn Song, and Jacob Steinhardt.
\newblock Measuring massive multitask language understanding.
\newblock \emph{arXiv preprint arXiv:2009.03300}, 2020.

\bibitem[Houlsby et~al.(2019)Houlsby, Giurgiu, Jastrzebski, Morrone, De~Laroussilhe, Gesmundo, Attariyan, and Gelly]{houlsby2019parameter}
Neil Houlsby, Andrei Giurgiu, Stanislaw Jastrzebski, Bruna Morrone, Quentin De~Laroussilhe, Andrea Gesmundo, Mona Attariyan, and Sylvain Gelly.
\newblock Parameter-efficient transfer learning for nlp.
\newblock In \emph{International conference on machine learning}, pages 2790--2799. PMLR, 2019.

\bibitem[Hu et~al.(2022)Hu, Shen, Wallis, Allen-Zhu, Li, Wang, Wang, Chen, et~al.]{Hu2022LoRA}
Edward~J Hu, Yelong Shen, Phillip Wallis, Zeyuan Allen-Zhu, Yuanzhi Li, Shean Wang, Lu~Wang, Weizhu Chen, et~al.
\newblock Lora: Low-rank adaptation of large language models.
\newblock \emph{ICLR}, 1\penalty0 (2):\penalty0 3, 2022.

\bibitem[Huan and Shun(2025)]{huan2025fine}
Muchen Huan and Jianhong Shun.
\newblock Fine-tuning transformers efficiently: A survey on lora and its impact.
\newblock 2025.

\bibitem[Li and Liang(2021)]{li2021prefix}
Xiang~Lisa Li and Percy Liang.
\newblock Prefix-tuning: Optimizing continuous prompts for generation.
\newblock \emph{arXiv preprint arXiv:2101.00190}, 2021.

\bibitem[Rumelhart et~al.(1986)Rumelhart, Hinton, and Williams]{rumelhart1986learning}
David~E Rumelhart, Geoffrey~E Hinton, and Ronald~J Williams.
\newblock Learning representations by back-propagating errors.
\newblock \emph{nature}, 323\penalty0 (6088):\penalty0 533--536, 1986.

\bibitem[Touvron et~al.(2023)Touvron, Martin, Stone, Albert, Almahairi, Babaei, Bashlykov, Batra, Bhargava, Bhosale, et~al.]{touvron2023llama}
Hugo Touvron, Louis Martin, Kevin Stone, Peter Albert, Amjad Almahairi, Yasmine Babaei, Nikolay Bashlykov, Soumya Batra, Prajjwal Bhargava, Shruti Bhosale, et~al.
\newblock Llama 2: Open foundation and fine-tuned chat models.
\newblock \emph{arXiv preprint arXiv:2307.09288}, 2023.

\bibitem[Wang et~al.(2023)Wang, Wu, Dabral, Zhang, Brown, Lu, Liu, Liang, Pang, Bendersky, et~al.]{wang2023non}
Yaqing Wang, Jialin Wu, Tanmaya Dabral, Jiageng Zhang, Geoff Brown, Chun-Ta Lu, Frederick Liu, Yi~Liang, Bo~Pang, Michael Bendersky, et~al.
\newblock Non-intrusive adaptation: Input-centric parameter-efficient fine-tuning for versatile multimodal modeling.
\newblock \emph{arXiv preprint arXiv:2310.12100}, 2023.

\bibitem[Woo et~al.(2025)Woo, Namkung, Lee, Jeong, Kim, and Jeon]{woo2025paca}
Sunghyeon Woo, Sol Namkung, Sunwoo Lee, Inho Jeong, Beomseok Kim, and Dongsuk Jeon.
\newblock Paca: Partial connection adaptation for efficient fine-tuning.
\newblock \emph{arXiv preprint arXiv:2503.01905}, 2025.

\bibitem[Yang et~al.(2024)Yang, Adamek, and Armour]{yang2024accurate}
Zeyu Yang, Karel Adamek, and Wesley Armour.
\newblock Accurate and convenient energy measurements for gpus: A detailed study of nvidia gpu’s built-in power sensor.
\newblock In \emph{SC24: International Conference for High Performance Computing, Networking, Storage and Analysis}, pages 1--17. IEEE, 2024.

\bibitem[Zaken et~al.(2021)Zaken, Ravfogel, and Goldberg]{zaken2021bitfit}
Elad~Ben Zaken, Shauli Ravfogel, and Yoav Goldberg.
\newblock Bitfit: Simple parameter-efficient fine-tuning for transformer-based masked language-models.
\newblock \emph{arXiv preprint arXiv:2106.10199}, 2021.

\bibitem[Zhang and Li(2021)]{zhang2021commentary}
Min Zhang and Juntao Li.
\newblock A commentary of gpt-3 in mit technology review 2021.
\newblock \emph{Fundamental Research}, 1\penalty0 (6):\penalty0 831--833, 2021.

\end{thebibliography}
\bibliographystyle{plainnat}

\end{document}